\documentclass[10pt,twocolumn,letterpaper]{article}

\usepackage{iccv}              

\definecolor{iccvblue}{rgb}{0.21,0.49,0.74}
\usepackage[pagebackref,breaklinks,colorlinks,allcolors=iccvblue]{hyperref}
\usepackage{pifont}
\colorlet{shadecolor}{gray!40}
\newcommand{\cmark}{\ding{51}}%
\newcommand{\xmark}{\color{shadecolor}\ding{55}}%

\def\paperID{*****} 
\def\confName{VisionDocs@ICCV}
\def\confYear{2025}

\title{CoSMo: A Multimodal Transformer for Page Stream Segmentation in Comic Books}

\author{Marc Serra Ortega$^{1}$ \\
{\tt\small marc.serrao@autonoma.cat}
\and
Emanuele Vivoli$^{1,2}$\\
{\tt\small evivoli@cvc.uab.cat}
\and 
Artemis Llabrés$^{1}$\\
{\tt\small allabres@cvc.uab.cat}
\and
Dimosthenis Karatzas$^{1}$\\
{\tt\small dimos@cvc.uab.cat}
\\
\\
Computer Vision Center and Universitat Autònoma de Barcelona$^{1}$\\
MICC, University of Florence, Italy$^{2}$
}

\begin{document}
\maketitle
\begin{abstract}
This paper introduces \textbf{CoSMo}, a novel multimodal Transformer for Page Stream Segmentation (PSS) in comic books, a critical task for automated content understanding, as it is a necessary first stage for many downstream tasks like character analysis, story indexing, or metadata enrichment. We formalize PSS for this unique medium and curate a new 20,800-page annotated dataset. CoSMo, developed in vision-only and multimodal variants, consistently outperforms traditional baselines and significantly larger general-purpose vision-language models across F1-Macro, Panoptic Quality, and stream-level metrics. Our findings highlight the dominance of visual features for comic PSS macro-structure, yet demonstrate multimodal benefits in resolving challenging ambiguities. CoSMo establishes a new state-of-the-art, paving the way for scalable comic book analysis. 
\end{abstract}

\section{Introduction}
Comics are a globally appreciated medium, enjoyed by readers of all ages---from Italian children collecting issues of \textit{Topolino}, to fans of American classics like \textit{Plastic Man}, to global audiences captivated by Japanese manga such as \textit{Naruto}. Beyond their inherent entertainment, comics serve as invaluable cultural artifacts, reflecting and chronicling societal values, historical moments, and evolving norms across different eras and geographies. Often published weekly or monthly, these books appear in anthology formats: short, self-contained stories from different authors and genres co-existing in a single issue, offering narrative continuity for ongoing plots while exposing readers to new styles and creators. These volumes are later recompiled into collected editions, but the original anthology structure remains a key publishing unit across decades and cultures.

\begin{figure}
    \centering
    \includegraphics[width=\linewidth]{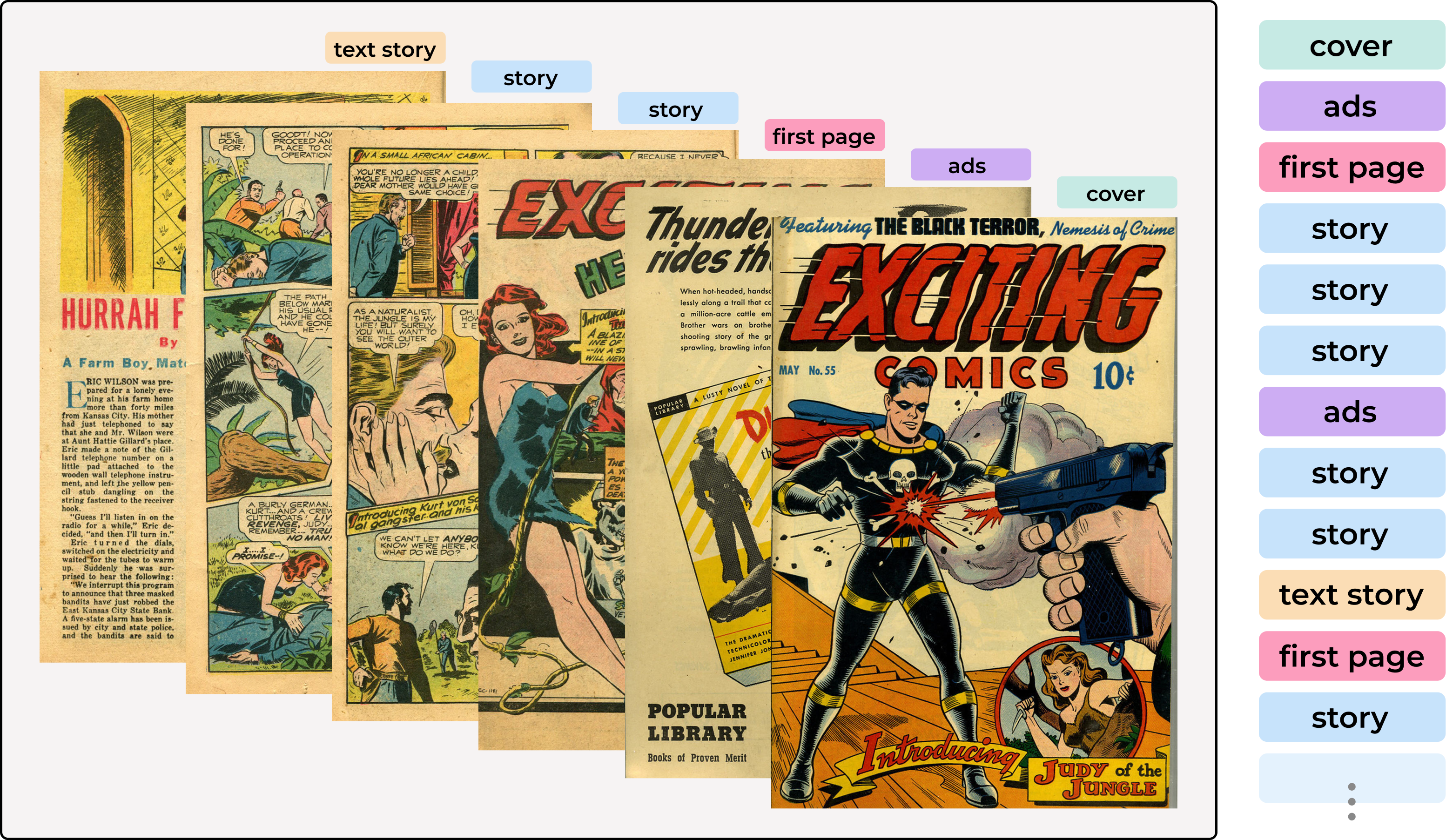}
     \caption{Page Stream Segmentation on a full comic book, splitting it into its semantic sections.}
    \label{fig:PSS}
\end{figure}

Much of this heritage has fortunately not been lost. Tens of thousands of historical comic books have been digitized and are now preserved in public repositories such as the \textit{Digital Comics Museum}\footnote{\url{https://digitalcomicmuseum.com/}} and \textit{Comic Book Plus}\footnote{\url{https://comicbookplus.com/}}. Crucially, these scanned books are enriched through manual annotation pipelines. Volunteers link each book to \textit{comics.org}\footnote{\url{https://www.comics.org/}}, the most comprehensive structured metadata archive in comics, which catalogs story titles, authors, page spans, characters, and editorial credits. These connections are made manually: a mediator retrieves metadata, identifies story boundaries, and tags each page accordingly. While this human effort has enabled large-scale archival access, it is unsustainable at scale. Automating this process is vital for preserving cultural heritage and enabling intelligent search and retrieval across massive comic corpora.

Comics pose unique challenges for computational analysis due to their combination of rich visuals, stylized typography, and non-linear layouts. Narrative content spans multiple modalities—image, text, and spatial structure—and varies widely across styles and cultures. These characteristics break standard document parsing assumptions, requiring models capable of multimodal reasoning.

We address the task of Page Stream Segmentation (PSS) in comics: the automatic division of scanned books into coherent sequences such as stories, advertisements, or text inserts (\cref{fig:PSS}). PSS is essential for enabling downstream tasks like story indexing or character analysis, yet remains largely unexplored in comics due to their structural complexity.

To tackle this, we introduce \textbf{CoSMo}, a Transformer-based architecture designed for comic PSS. We curated a dataset of 430 manually annotated books from public archives, correcting inconsistencies in available metadata. CoSMo consistently outperforms baselines and even large multimodal language models, despite being lightweight, demonstrating its effectiveness across a range of evaluation metrics and settings.

Our main contributions are:
\begin{itemize}
    \item We formalized the PSS task for comic books and created a high-quality annotated dataset, meticulously aligned with \textit{comics.org} metadata.
    \item We develop CoSMo, a modular Transformer encoder available in a lightweight vision-only configuration and a robust multimodal variant.
    \item We perform extensive experiments and ablations to rigorously evaluate the impact of diverse representations and benchmark against a wide array of baselines, showcasing CoSMo's superior performance compared to traditional methods and zero-shot multimodal LLMs.
\end{itemize}

All code, data, and annotations are available in GitHub\footnote{\url{https://github.com/mserra0/CoSMo-ComicsPSS}} to empower the comics community to automatically generate accurate annotations for their extensive corpus of comic books, a task our method performs with an impressive error rate of approximately 1\%.

\section{Problem Definition}

Page Stream Segmentation (PSS) for comic books is fundamentally defined as the task of identifying semantic boundaries to form coherent and meaningful groups of subsequent pages within a sequential stream of scanned or digitized comic pages. In this work, we reframe this as a single-page classification problem, which involves assigning a semantic label to each page. Given a document stream represented as an ordered sequence of $n$ pages: $\mathcal{S} = \{p_1, p_2, \dots, p_n\}$, the goal is to learn a segmentation function $f_\theta: \mathcal{S} \rightarrow \{y_1, y_2, \dots, y_n\}$, where each $y_i \in \mathcal{Y}$ is the class label associated with page $p_i$, and $\mathcal{Y}$ is the set of predefined semantic categories: $\mathcal{Y} = \{\text{Cover}, \text{Advertisement}, \text{Story}, \text{Text Story}\}$.

This problem is cast as a multiclass sequence labeling task, where each prediction may depend not only on the visual and textual content of the current page but also on the broader narrative context across the page stream. The model must integrate multimodal cues---visual layout, stylistic features, and text---across the sequence to produce accurate and coherent segmentations. This formulation grounds PSS in a structured learning framework and motivates the specific research objectives described below.

\section{State of the Art}

\textbf{Comic Books.} 
The fast-growing literature on computational comics understanding has recently been condensed by Vivoli et al.~\cite{vivoli2025missingpiecevisionlanguage}, introducing the \textit{``Layer of Comics Understanding (LoCU)''} framework to situate vision–and–language tasks along the axes of input/output modality and cognitive complexity. Their survey exposes three systemic bottlenecks---limited open datasets, weak reproducibility, and an over-reliance on single-page, low-level vision settings. In response, they released two complementary resources: the \textit{Comics Dataset Framework (CDF)} \cite{vivoli2024comicsdatasetsframeworkmix}, which unifies annotation schemas to facilitate reproducibility, and \textit{CoMix} \cite{vivoli2024comixcomprehensivebenchmarkmultitask}, a multitask benchmark that probes previously unexplored single-page capabilities such as character naming and dialogue generation. Building on this foundation, the \textit{Magi} series pushes analysis into higher \textit{``LoCU layers''}, advancing detection, character re-identification, and narrative generation for manga pages \cite{sachdeva2024mangawhispererautomaticallygenerating,Sachdeva_2024_ACCV,sachdeva2025panelsprosegeneratingliterary}. Beyond the single-page regime, multimodal reasoning across panel sequences has been explored via the text-cloze task \cite{Iyyer2016TheAM,vivoli2024multimodaltransformercomicstextcloze}. Yet, to date, no method ingests an entire volume---crucial for book-level understanding, metadata extraction, or coherent page grouping. Closing this gap is the central objective of the present work.

\textbf{Page Stream Segmentation.}
PSS is a fundamental task in document analysis, aimed at segmenting a continuous sequence of pages into coherent document units (e.g., individual issues, stories, or sections). This task has been extensively explored in domains such as banking, business, and legal workflows. Early methods heavily relied on OCR to extract text-based features, which were then passed to traditional machine learning models~\cite{demirtas2022semanticparsinginterpagerelations, wiedemann2019pagestreamsegmentationconvolutional, tapthisfolder, documentClassificationMailroom}.

Over time, these approaches evolved into multimodal systems combining CNNs for visual input and RNNs for textual input, enabling the models to learn richer representations of page structure and semantics~\cite{demirtas2022semanticparsinginterpagerelations}. More recently, Transformer-based architectures have been applied in PSS too, revealing that decoder-only large language models (LLMs), fine-tuned with parameter-efficient techniques, can leverage textual features alone to surpass smaller, dedicated multimodal PSS approaches~\cite{heidenreich2024largelanguagemodelspage}.

Standardisation has followed suit. OpenPSS \cite{openpss} unifies earlier task formulations---spanning Tab This Folder \cite{tapthisfolder} and the panoptic-segmentation literature \cite{kirillov2019panopticsegmentation}---into a single benchmark with harmonised document- and stream-level metrics, enabling apples-to-apples comparison across domains.

Despite this progress, comics remain conspicuously absent from the dedicated PSS corpus.

\section{Methodology}
\label{sec:methodology}
In this section, we introduce \textbf{CoSMo} (Comic Stream Modeling), our novel transformer-encoder model for robust Page Stream Segmentation (PSS) in comic books. CoSMo's design explicitly tackles the unique multimodal challenges inherent in comic book documents by integrating both visual and textual information within a sequential context. To offer both comprehensive and cost-effective solutions, CoSMo is developed in two primary variants: a \textbf{multimodal model} that leverages visual and textual cues, and a \textbf{vision-only model} optimized for scenarios with textual processing limitations. Both variants share a common architectural inspiration from the successful encoder-only Transformer design.

\subsection{\textit{CoSMo}}

\subsubsection{Multimodal Architecture}
\label{sec:CoSMoMultimodal}
\begin{figure}[t!]
\centering
\includegraphics[width=\linewidth]{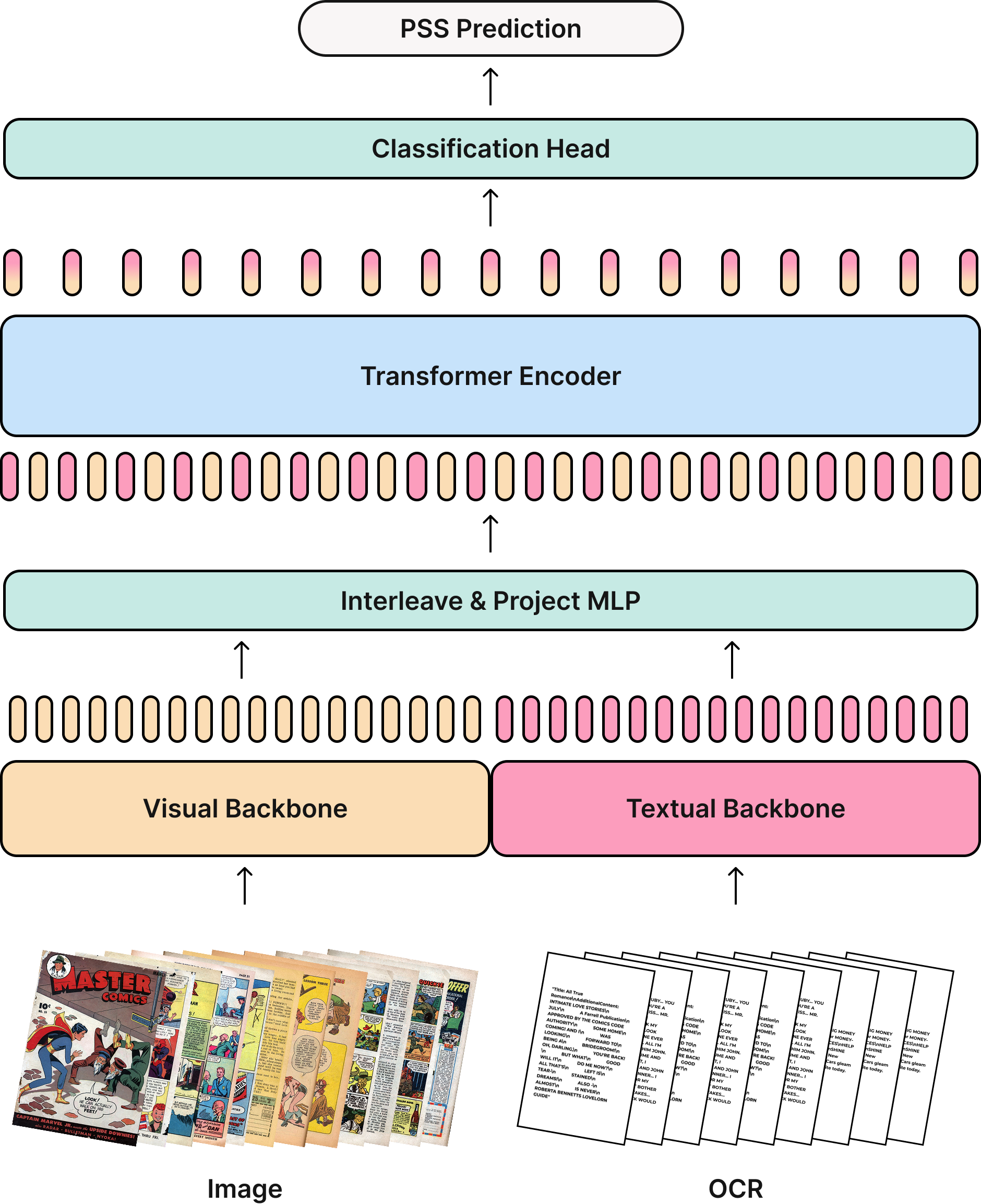}  
\caption{Multimodal CoSMo model architecture.}
\label{fig:MultimodalCoSMo}
\end{figure}

The \textit{CoSMo} \textbf{Multimodal} architecture, as shown in \cref{fig:MultimodalCoSMo}, integrates visual features with contextualized textual embeddings, feeding them into a Transformer encoder for sequence processing.

\textbf{Visual Feature Extraction:} Visual features are obtained from each page using a frozen \textbf{SigLIP} backbone. SigLIP is chosen for its strong performance in general image representation, providing robust visual embeddings.

\textbf{Textual Feature Extraction:}\label{par:TextExtractor} We extracted rich, contextualized OCR from Qwen2.5-VL-32B, which provided a comprehensive textual understanding by identifying both the reading order and distinct text types, including \textit{Titles}, \textit{Panels}, and \textit{Content Text}, particularly relevant for pages like \textit{Text-Stories} and \textit{Advertisements}. This structured OCR output was then embedded using Qwen3Embedding-0.6B \footnote{\url{https://hf.co/Qwen/Qwen3-Embedding-0.6B}}\cite{zhang2025qwen3embeddingadvancingtext}. This embedding model, specifically tailored for textual representation, effectively leverages the advanced semantic understanding capabilities of large language models.

\textbf{Feature Projection and Interleaving:}
Before applying the multi-token arrangement, both visual and textual feature vectors are independently projected to a shared dimensionality of 768 using a three-layer MLP. This projection module incorporates GELU activations, dropout (rate = 0.4), and layer normalization to ensure stable and expressive feature transformations. After projection, the resulting embeddings are normalized to maintain consistent feature scaling across modalities. For interleaving, we construct a dual-token representation per page---one for the visual modality and one for the textual modality---allowing the Transformer encoder to jointly attend over both representations in sequence. This structure enables the model to capture complementary signals and cross-modal dependencies effectively.

\textbf{Transformer Encoder:}
The page-level multimodal representations are processed by a encoder-only Transformer, which models contextual dependencies across the entire comic book sequence. The encoder consists of 4 Transformer layers with 4 self-attention heads per layer. It uses a hidden size of 256, an input embedding size of 768, and an intermediate feedforward dimension of 3072 (i.e., 4× the hidden size). To regularize training, a dropout rate of 0.4 is applied throughout the network. We adopt absolute positional encoding to preserve the ordering of pages within the book, which is essential for learning narrative flow and structural transitions.
\textbf{Classification Head:}
From the dual-token sequence, only the second token corresponding to each page is forwarded to the classification head. This head consists of a three-layer MLP that includes GELU activations, layer normalization, and a dropout rate of 0.4. It maps the contextualized representation of each page to one of the predefined semantic categories (e.g., story, cover, advertisement), enabling fine-grained classification across the comic stream.

\subsubsection{Vision-Only Architecture}
The \textit{CoSMo} \textbf{Vision-Only} model offers a computationally efficient alternative to its multimodal counterpart by relying exclusively on visual information. This variant retains the same core Transformer encoder and classification head but omits the OCR extraction, textual embedding, and the interleaving components---substantially reducing model complexity and the need for OCR. The motivation behind this design stems from two key observations: first, high-quality OCR in comics is often computationally expensive due to complex layouts and stylized fonts; second, as shown in our experiments, visual features alone already capture much of the structural and semantic information required for accurate segmentation. Consequently, the Vision-Only CoSMo model presents a highly competitive and cost-effective solution for Page Stream Segmentation, achieving strong results with minimal trade-off in performance.

\section{Dataset}

Our dataset consists of 430 \textit{classic} comic books sourced from the Digital Comic Museum (DCM)\footnote{\url{https://digitalcomicmuseum.com/}}, a public-domain archive of \textit{Golden Age comics}. We manually annotated these books and performed thorough quality checks to ensure accuracy. This collection contains over 20,800 pages, ensuring stylistic and structural diversity while supporting deep model training.

Each page is labeled with one of five semantic classes: \textit{Cover}, \textit{Advertisement}, \textit{Text Story}, \textit{Story}, or \textit{First-Page}---the latter being a derived label marking the first page in a narrative block. These classes reflect common structural components of vintage comic books.

Comics typically begin with a \textit{Cover}, followed by a mix of \textit{Advertisements}, \textit{Text Stories}, and one or more \textit{Story} segments. Transitional pages often separate story blocks, but this structure varies: some books feature continuous narratives, others have only one or two categories.

\begin{figure}[t!]
\centering
\includegraphics[width=1.0\linewidth]{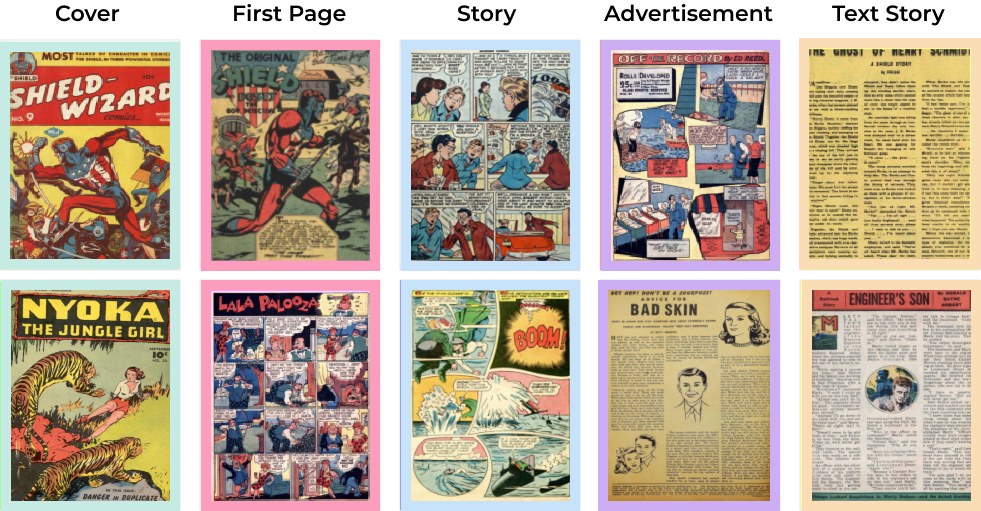}
\caption{Examples of annotated comic book page types used for multiclass page stream segmentation. \textit{First-Page} is a derived label marking story segment beginnings.}
\label{fig:exampleClasses}
\end{figure}

\noindent The dataset poses several challenges:

\textbf{Intra-class diversity:} Pages in the same class (e.g., \textit{Advertisement}) differ greatly in layout and style as illustrates in \cref{fig:exampleClasses}.

\textbf{Inter-class similarity:} \textit{Text Stories} and text-heavy \textit{Advertisements} are often visually similar. Likewise, \textit{First-Page} and \textit{Story} pages can be hard to distinguish without narrative context.

\textbf{Class imbalance:} \textit{Story} pages dominate the dataset (71\%), while other classes are underrepresented: 2.4\% \textit{Cover}, 8.8\% \textit{Advertisement}, 13.4\% \textit{First-page} and 4.2\% \textit{Text-story}.

\section{Experiments}

This section outlines the experimental framework used to evaluate CoSMo and several baselines for the PSS task. We describe both single-page and multi-page modeling paradigms, including variations in input modalities and backbone architectures, to systematically explore the challenges of segmenting comic book pages.

\subsection{Baselines}

\textbf{Handcrafted Detection Features with Traditional ML.}
We began by extracting semantic layout elements---characters, faces, panels, text blocks---from each page using the object detection module from MAGI.

Building upon these detections, 20 handcrafted features were extracted for each page. These features, designed to characterize content density, spatial layout, and element distributions, were crucial for discerning structural transitions within the sequence. We then utilized these features to train an XGBoost model.

\textbf{Pretrained Visual Backbones}
To assess visual representation quality, we tested CLIP and SigLIP under two regimes:

\textbf{(i)} Zero-shot classification via prompt-based inference, and  
\textbf{(ii)} Linear probing using a lightweight MLP trained on frozen visual embeddings.

These experiments aimed to evaluate both the transferability of vision–language models to the comic domain and their capacity for fine-grained semantic separation. Prompt tuning was used to improve zero-shot alignment with class labels. The linear probe setup allowed direct comparison of learned representations without full fine-tuning.

\textbf{Zero-Shot LLM Experiments}
To evaluate large-scale instruction-tuned models in a zero-shot setting, we tested Qwen2.5-VL-32B \footnote{\url{https://hf.co/Qwen/Qwen2.5-VL-32B-Instruct}}\cite{bai2025qwen25vltechnicalreport} using two prompt formats:  
(i) a CLS-focused prompt focused on providing a short caption followed by a direct classification of the image, and  
(ii) an OCR-focused prompt that added the OCR extraction, encouraging text interpretation before classification.  
This helped explore how prompt design affects segmentation accuracy and whether massive models can generalize to comic PSS without task-specific fine-tuning.

\subsection{Multi-page CoSMo}

Our base model, CoSMo, operates on frozen visual embeddings from backbone models and serves as the foundation for all multi-page experiments.

\textbf{Visual Backbone Ablation Study.} We evaluated four pretrained visual backbones---CLIP, SigLIP, SigLIP2, and DINOv2---within the CoSMo architecture. These models differ in training paradigms and input resolution.

\textbf{Detection Feature Fusion.} To incorporate structural layout cues, we fused handcrafted features into the best visual-only CoSMo model (SigLIP backbone). Features were projected, normalized, concatenated with visual embeddings, and passed through an MLP before being processed by the Transformer encoder. This tested whether explicitly injected layout signals improve segmentation of ambiguous classes.

\textbf{Text-Only Modeling.}
In our Text-Only Modeling approach, textual features, obtained via the strategy detailed in \cref{par:TextExtractor}, constituted the sole input to CoSMo. This configuration allowed us to assess the efficacy of OCR-based representations in distinguishing fine-grained page roles, especially under conditions where visual cues are ambiguous.

\textbf{Multimodal CoSMo} Finally, the full multimodal CoSMo variant was evaluated by integrating SigLIP visual embeddings with Qwen-derived OCR embeddings, these modalities were combined using two distinct strategies---fused, where projected embeddings were concatenated and passed through a two-layer MLP, and multitoken, where each embedding was treated as a separate input token---before being encoded through the same Transformer backbone.

\subsection{Evaluation Metrics}

To rigorously evaluate the performance of our Page Stream Segmentation (PSS) models, we employ a combination of page-level and document-level metrics tailored to both single-page and multi-page modeling scenarios, as well as providing standardized metrics in the PSS field \cite{openpss}. These metrics assess not only classification accuracy but also the quality of sequential predictions, which are crucial in structured documents such as comic books.

\subsubsection{Single-Page Level Evaluation}

For single-page modeling, the classification task is framed as a multi-class problem with significant class imbalance. To address this, we adopt the \textbf{Macro-averaged F1 score (F1-Macro)} as our primary evaluation metric. This metric is consistently used across all experiments as our primary metric.

The F1-Macro score is defined as the harmonic mean of precision and recall, calculated for each class and then averaged across all classes. The formula for F1-Macro is:
\begin{equation}
F1_{\text{Macro}} = \frac{1}{N} \sum_{i=1}^{N} \frac{2\text{P}_i\text{R}_i}{\text{P}_i + \text{R}_i}
\end{equation}
where: $N$ is the number of classes, $\text{P}_i$ and $\text{R}_i$ are the precision and recall for class $i$, respectively. This formulation ensures fair evaluation across rare and frequent classes. Additionally, we analyze the confusion matrix to gain insights into class-specific prediction strengths and weaknesses.

\begin{table*}[t!]
\centering
\caption{F1-Macro and Accuracy for multiclass single-page classification under different training setups.}
\label{tab:single-page-multiclass}
\begin{tabular}{@{}lcccc@{}}
\toprule
\textbf{Model}       & \textbf{Modality} & \textbf{Setup}       & \textbf{F1-Macro $\uparrow$} & \textbf{Accuracy $\uparrow$} \\ \midrule
XGBoost              & Detection         & End-to-End           & 83.50             & 90.76          \\
CLIP                 & Vision + Prompt   & Zero-Shot            & 61.90             & 80.30        \\
SigLIP               & Vision + Prompt   & Zero-Shot            & 62.70             & 84.90        \\
CLIP + MLP           & Vision            & Linear Probe         & 80.64            & 90.59       
\\
SigLIP + MLP         & Vision            & Linear Probe         & 89.92             & 95.81        \\
Qwen (CLS Prompt)    & Vision + Prompt   & Zero-Shot            & 87.12             & 92.25        \\
Qwen (OCR Prompt)    & Vision + Prompt   & Zero-Shot            & 88.26             & 93.84  
\\
CoSMo                & Vision            & End-to-End           & \underline{93.46}             & \underline{96.16}
\\
CoSMo                & Multimodal        & End-to-End           & \textbf{97.82}             & \textbf{98.41}
\\ \bottomrule
\end{tabular}
\end{table*}

\subsubsection{Multi-Page Level Evaluation}

The evaluation of multi-page models requires sequence-aware metrics that assess not only the correctness of individual page classifications but also the coherence and accuracy of entire document structures.

\textbf{Document-Level Metrics:} We implement \textbf{Document-Level F1}, as proposed in prior works \cite{openpss,tapthisfolder}, which computes F1 scores at the level of contiguous semantic segments (e.g., stories or advertisement blocks), rather than individual pages. Furthermore, inspired by the Panoptic Segmentation task in computer vision \cite{kirillov2019panopticsegmentation}, we adopt a unified metric called \textbf{Panoptic Quality (PQ)}, which measures both recognition quality and segmentation quality: 

\begin{equation}
\text{PQ} = \text{DocF1} \times \text{SQ}
\end{equation}

Where: \textbf{DocF1} is the harmonic mean of segment-level precision and recall, and \textbf{Segmentation Quality (SQ)} is defined as: 
\begin{equation}
    \text{SQ} = \frac{1}{|TP|} \sum_{(p, g) \in TP} \frac{|p \cap g|}{|p \cup g|}
\end{equation}
Segments are considered matched when their Intersection-over-Union (IoU) exceeds a certain threshold, 0.5 in our case. Based on these matched segments, we compute standard document-level metrics, including Precision, Recall, and F1, derived from true positives (TP), false positives (FP), and false negatives (FN).

\textbf{Stream-Level Metrics:} To evaluate the model's utility in real-world annotation workflows, we implement the user-centric metric Minimum Number of Drags and Drops  (MnDD) presented in \cite{tapthisfolder}. This metric measures the minimal number of operations needed to transform the predicted segmentation into the ground truth using a graphical interface. It is defined as:

\begin{equation}
\text{MnDD} = N - \sum_{i,j} \max_i |G_i \cap P_j|
\end{equation}

where $G_i$ and $P_j$ are ground truth and predicted segments, respectively.

\subsection{Training Protocol}

CoSMo models are trained using a cost-sensitive Cross-Entropy loss function, where each class's contribution is weighted inversely to its frequency to mitigate class imbalance, optimizing for multiclass sequence labeling. Training is performed with a learning rate of $1\cdot10^{-6}$ and employs Early Stopping to prevent overfitting, with monitoring conducted on an L40s GPU.

\section{Results}
This section presents the empirical evaluation of our proposed CoSMo models on the PSS task. Subsequently, we detail results from our ablation studies, examining the influence of backbones, fusion strategies, and input modalities. Finally, we present qualitative results and discuss key insights.

\subsection{Single-Page Baselines}
To understand the capacity of isolated page-level information for inferring semantic roles, we first evaluated various single-page classification models, using F1-Macro as the primary metric due to dataset imbalance. The full results are provided in \cref{tab:single-page-multiclass}.

Experiments on single-page classification revealed several key insights. The \textbf{XGBoost classifier}, using handcrafted features, achieved reasonable overall performance but struggled with fine-grained distinctions---particularly for the critical \textit{First Page} class---due to its reliance on layout-only cues.

\textbf{Zero-shot classification} using \textbf{CLIP} and \textbf{SigLIP} showed strong performance on broad categories like \textit{Cover}, \textit{Advertisement}, and \textit{Story}. However, both models failed to distinguish nuanced classes such as \textit{First Page} and \textit{Text Story}, indicating a lack of domain-specific sensitivity in purely contrastive vision–language embeddings.

\textbf{Linear probing} on frozen visual embeddings led to significant improvements, particularly for challenging categories. Notably, SigLIP surpassed CLIP in this setting, achieving high accuracy without the need for end-to-end fine-tuning---highlighting the quality and structure of its learned representation space.

\textbf{Zero-shot evaluation with Qwen2.5-VL-32B} delivered strong results, especially when using OCR-style prompting. This approach improved performance on context-dependent classes by encouraging the model to ``read" the page. However, despite its capabilities, Qwen2.5-VL-32B lagged behind the lighter SigLIP Linear Probe in accuracy and demanded far greater computational resources.

\textbf{CoSMo} was also evaluated in a single-page setting, despite being primarily trained on full page sequences, to assess its robustness without explicit sequential context. In this configuration, the Vision-Only variant exhibited a significant drop in its ability to detect \textit{First Page} transitions, with accuracy for this class falling from 96\% (multi-page setting) to 77\%. In stark contrast, the Multimodal CoSMo variant demonstrated remarkable resilience; its \textit{First Page} accuracy only decreased from 97\% (multi-page) to 92\% in the single-page setting. These results offer two critical insights: First, the multimodal approach's benefit, particularly in challenging single-page scenarios, stems from its ability to extract crucial context from textual features. Second, both models significantly benefit from the contextualized Transformer architecture's capacity to capture implicit relations and order, even when presented with isolated pages.

\subsection{Multi-Page Results}

\begin{table*}[t!]
\centering
\caption{Performance summary of CoSMo variants across modalities and integration strategies.}
\label{tab:cosmo-results-summary}
\begin{tabular}{@{}lcccc|cccc@{}}
\hline
 & \textbf{Vision}        & \textbf{Textual}    & \textbf{Detection} & \textbf{Fusion} & \textbf{F1 Macro $\uparrow$} & \textbf{Acc $\uparrow$} & \textbf{PQ $\uparrow$} & \textbf{MnDD $\downarrow$} \\ \hline
Vision-Only & \cmark & \xmark & \xmark & \xmark         & \underline{97.30} & \underline{98.46} & 94.50 & \underline{0.632}  \\
Text-Only   & \xmark & \cmark & \xmark & \xmark         & 87.92 & 88.90 & 70.30 & 6.322 \\
Vision+Detection   & \cmark & \xmark & \cmark & \cmark  & 96.12 & 97.63 & 92.68 & 1.069\\
Multimodal  & \cmark & \cmark & \xmark & \xmark         & \textbf{98.10} & \textbf{98.65} & \textbf{95.08} & \textbf{0.437} \\
Multimodal  & \cmark & \cmark & \xmark & \cmark         & 96.82 & 98.21 & \underline{94.75} & 0.667 \\ \hline
\end{tabular}
\end{table*}

We now evaluate whether modeling sequential context across pages improves the segmentation of structurally important classes in comics. \cref{tab:cosmo-results-summary} summarizes the performance of various CoSMo architecture variants, progressively incorporating different modalities and design choices.

\textbf{Detection Feature Fusion.}
Incorporating detection features into CoSMo led to competitive performance across metrics. However, results remained slightly below the vision-only baseline, suggesting that while these features provide useful layout biases, their benefit may be limited by extraction inconsistencies. In particular, the MAGI detector---originally trained on Manga---may not generalize reliably across diverse comic styles, introducing noise when combined with rich visual embeddings.

\textbf{Textual-Only CoSMo.}
Despite the absence of visual input, the textual-only model achieved a respectable F1-Macro score. However, it performed poorly on structural metrics such as PQ and MnDD, largely due to frequent misclassification of \textit{First Page} transitions. These errors stem from the OCR model's difficulty in extracting stylized or decorative title elements critical for story boundary detection. While text provides useful semantic cues, it lacks the spatial and stylistic grounding required for precise segmentation, reinforcing its role as a complementary modality rather than a standalone input.

\textbf{Visual-Only CoSMo.}
The vision-only CoSMo model, built on the SigLIP backbone, delivers strong performance across all metrics, with an F1-Macro of 97.30 and a Panoptic Quality (PQ) of 94.50. Its low stream-level error (MnDD of 0.632) confirms that visual features alone are highly effective at modeling both semantic and structural aspects of comic narratives.

\textbf{Multimodal CoSMo.}
Incorporating textual features yields the best overall performance, reaching a 98.10 F1-Macro, 95.08 PQ, and the lowest MnDD of 0.437. These gains suggest that text provides complementary cues that help resolve subtle ambiguities, especially in structurally complex or visually unconventional pages. However, improvements over the vision-only model remain relatively modest considering the added complexity of OCR and fusion. 

\subsection{Ablation Study}
\begin{table}[t!]
\centering
\caption{Multi-page modeling results for CoSMo Base with different visual backbones.}
\label{tab:multi-backbone-results}
\resizebox{\columnwidth}{!}{%
\begin{tabular}{@{}lllll@{}}
\toprule
\textbf{Backbone} & \textbf{F1-Macro $\uparrow$} & \textbf{Acc $\uparrow$} & \textbf{PQ $\uparrow$}    & \textbf{MnDD $\downarrow$}  \\ \midrule
SigLIP            & \textbf{97.30}    &  \textbf{98.46}  & \textbf{94.50} & \textbf{0.632} \\
SigLIP2           & 96.27             &  97.77  & 93.66          & 0.7931          \\
CLIP              & 91.44             &  91.46  & 76.72          & 4.6552          \\
DINOv2            & 88.62             &  91.07  & 75.13          & 4.8391          \\ \bottomrule
\end{tabular}%
}
\end{table}

To identify the best visual backbone for CoSMo, we conducted an ablation study with various pretrained vision encoders, shown in \cref{tab:multi-backbone-results}.

SigLIP and SigLIP2 performed strongly, with SigLIP achieving the best results in all metrics. Despite its larger size, SigLIP2 slightly underperformed, suggesting that architecture alone doesn’t guarantee better segmentation without task-specific alignment.

CLIP and DINOv2, while effective in classification, struggled with segmentation---especially on the critical \textit{First Page} class---highlighting their limitations in capturing the stylistic and structural nuances of comics.

Across models, \textit{First Page} remained the hardest class due to its similarity to \textit{Story} pages. SigLIP handled this best, validating our hypothesis that Transformer-based sequence modeling enhances performance on structurally ambiguous cases. These findings position SigLIP as the best visual backbone for the CoSMo model, and hence it is the one used in all experiments.

\subsection{Qualitative Results}

\begin{figure*}[t!]
\centering
\includegraphics[width=0.8\linewidth]{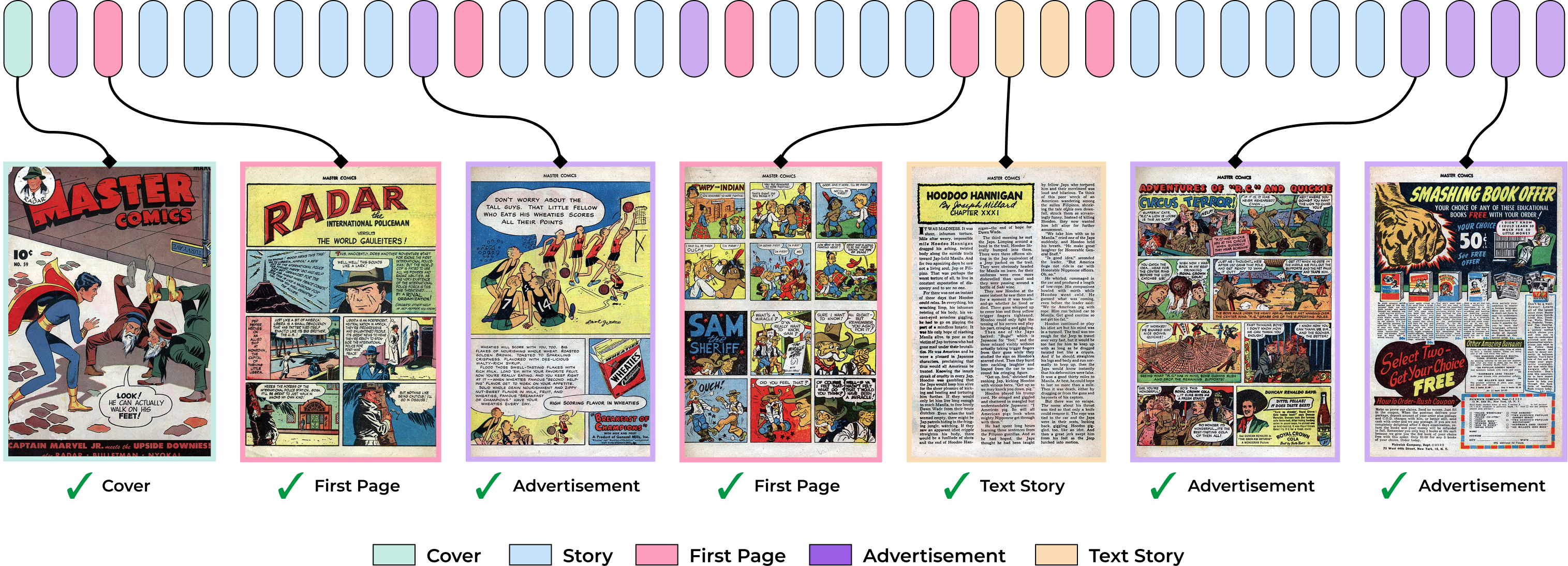} 
\caption{Example of a comic page sequence showcasing robust and accurate page stream segmentation by both CoSMo Vision and CoSMo Multimodal.}
\label{fig:GGexample}
\end{figure*}

\begin{figure*}[t!]
  \centering
  \begin{minipage}[t]{0.49\textwidth}
    \centering
    \includegraphics[width=\linewidth]{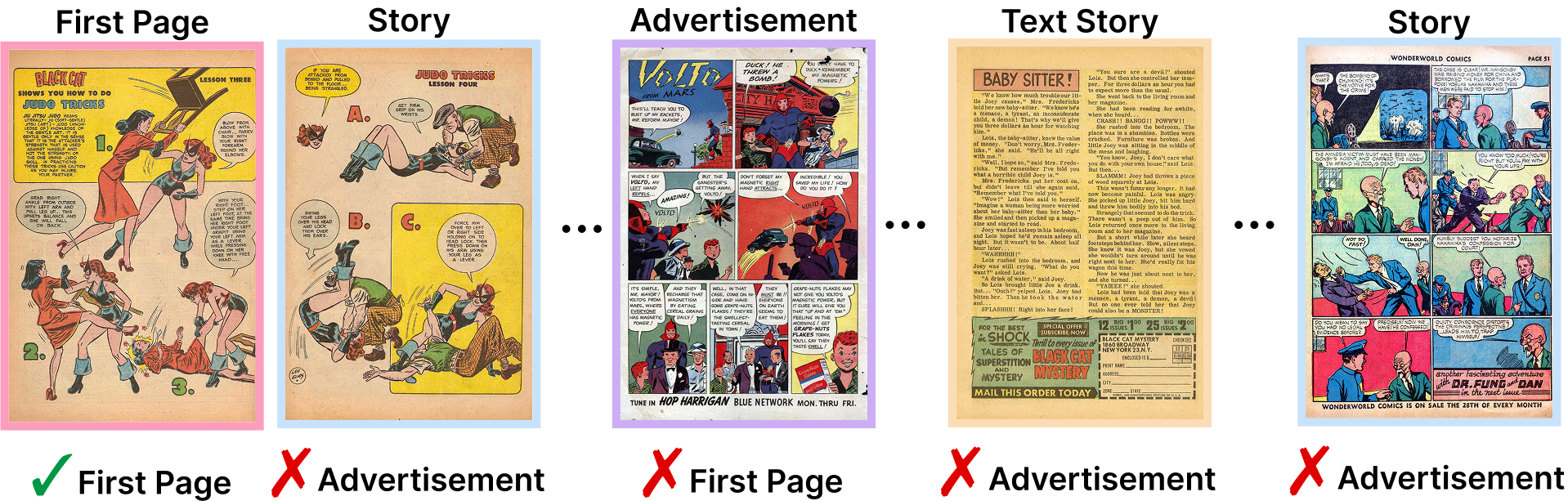}
    \caption{Examples of pages correctly classified by CoSMo Multimodal and misclassified by CoSMo Vision, highlighting the critical role of textual features in challenging scenarios.}
    \label{fig:ExampleGB}
  \end{minipage}%
  \hfill
  \begin{minipage}[t]{0.49\textwidth}
    \centering
    \includegraphics[width=\linewidth]{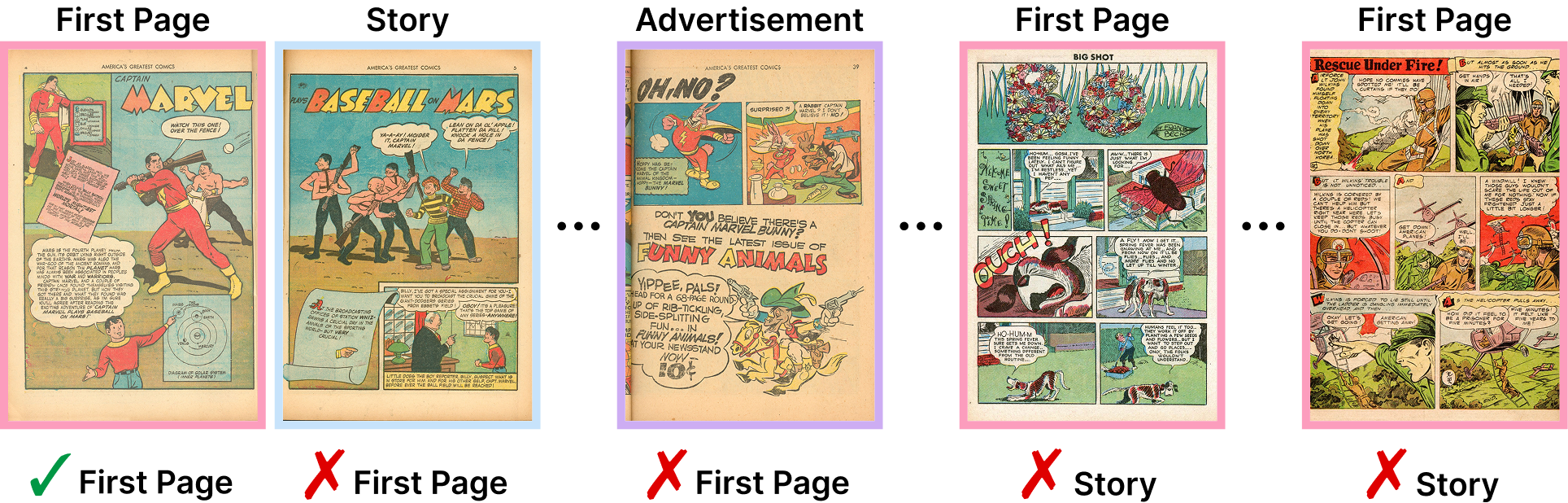}
    \caption{Pages misclassified by both CoSMo variants, illustrating common error patterns that challenge both vision-only and multimodal models.}
    \label{fig:ExampleBB}
  \end{minipage}
\end{figure*}


To complement the quantitative analysis, this section provides visual examples of CoSMo's performance, highlighting its ability to accurately segment comic page streams, including common challenges, and demonstrate the nuanced contributions of multimodal information. 

\textbf{Successful Segmentation Examples.} \cref{fig:GGexample} illustrates a full comic book stream that both CoSMo Vision-Only and CoSMo Multimodal successfully segmented. This example was carefully chosen for its inherent ambiguity, showcasing both standard and challenging page types, including a regular \textit{Cover}, a typical \textit{First-page }, a difficult \textit{Advertisement} resembling a \textit{Story}, a complex \textit{First-page} with multiple small titles, a regular \textit{Text-story}, another challenging \textit{Advertisement} easily mistaken for a \textit{First-page} or \textit{Story}, and a standard \textit{Advertisement}. This success demonstrates their robust capacity to learn complex visual and sequential patterns. 

\textbf{Advantages of the Multimodal Approach.} While the CoSMo Vision-Only model performs well overall, it struggles with certain ambiguous cases due to the diverse visual styles of comic pages. \cref{fig:ExampleGB} illustrates scenarios where adding textual features resolves these ambiguities. The multimodal model correctly distinguishes \textit{First Page} from \textit{Advertisement}, detects atypical \textit{Advertisements} with misleading layouts, and accurately classifies mixed-content pages like \textit{Text Stories}. These examples highlight how textual cues enhance fine-grained semantic understanding, especially in cases where visual information alone is insufficient.

\textbf{Analysis of Persistent Challenges.} Despite strong overall performance, both CoSMo variants struggle with certain edge cases, as shown in \cref{fig:ExampleBB}. Common failure modes include misclassifying non-initial \textit{Story} pages as \textit{First Page} due to local title cues without considering preceding context. Similarly, pages combining visual storytelling and ad-like content often confuse the model. Legitimate \textit{First Page} examples with atypical layouts or weak title cues are also frequently mislabeled as \textit{Story}. These errors underscore the difficulty of modeling long-range dependencies and handling visual and stylistic variability---key areas for future improvement.

\section{Conclusions}
We introduced CoSMo, a multimodal Transformer for Page Stream Segmentation (PSS) in comic books, alongside a curated annotated dataset. CoSMo achieved state-of-the-art results, with the vision-only variant (SigLIP backbone) proving highly effective and efficient. While visual cues dominate, adding textual features in the CoSMo Multimodal, further improved performance in ambiguous cases, though at the cost of increased computational load and OCR dependency. CoSMo also outperformed much larger general-purpose models, highlighting the value of task-specific architectures for structured document understanding and marking a key step toward layered comic analysis in the LoCU framework.


\section*{Acknowledgements}
This paper has been supported by the Consolidated Research Group 2021 SGR 01559 from the Research and University Department of the Catalan Government, and by project PID2023-146426NB-100 funded by MCIU/AEI/10.13039/501100011033 and FSE+. With the support of the FI SDUR predoctoral grant from the Generalitat de Catalunya and co-financing by the FSE+ (2024FISDU\_00095).

{
    \small
    \bibliographystyle{ieeenat_fullname}
    \bibliography{main}
}

\end{document}